\documentclass[11pt,a4paper]{article}
\usepackage[hyperref]{emnlp2020}
\usepackage{times}
\usepackage{latexsym}

\usepackage{setspace}
\usepackage{microtype}
\usepackage{graphicx}
\usepackage{amsmath}
\usepackage{bbold}
\usepackage{dsfont}

\usepackage{graphicx}

\makeatletter
\newcommand*\bigcdot{\mathpalette\bigcdot@{.5}}
\newcommand*\bigcdot@[2]{\mathbin{\vcenter{\hbox{\scalebox{#2}{$\m@th#1\bullet$}}}}}
\makeatother

\aclfinalcopy 
\def\aclpaperid{2758} 

\title{\textit{Reading Between the Lines:} Exploring Infilling in Visual Narratives}

\author{
Khyathi Raghavi Chandu  \qquad  Ruo-Ping Dong  \qquad  Alan W  Black\\
\\
Language Technologies Institute, Carnegie Mellon University \\
\tt{ \{kchandu, awb\}@cs.cmu.edu, {ruopingd}@alumni.cmu.edu }
}

\date{}

\begin{document}
\maketitle
\begin{abstract}
Generating long form narratives such as stories and procedures from multiple modalities has been a long standing dream for artificial intelligence. In this regard, there is often crucial subtext that is derived from the surrounding contexts. The general seq2seq training methods render the models shorthanded while attempting to bridge the gap between these neighbouring contexts. In this paper, we tackle this problem by using \textit{infilling} techniques involving prediction of missing steps in a narrative while generating textual descriptions from a sequence of images. We also present a new large scale \textit{visual procedure telling} (ViPT) dataset with a total of 46,200 procedures and around 340k pairwise images and textual descriptions that is rich in such contextual dependencies. Generating steps using infilling technique demonstrates the effectiveness in visual procedures with more coherent texts. We conclusively show a METEOR score of 27.51 on procedures which is higher than the state-of-the-art on visual storytelling. We also demonstrate the effects of interposing new text with missing images during inference. The code and the dataset  will be publicly available at \href{https://visual-narratives.github.io/Visual-Narratives/}{https://visual-narratives.github.io/Visual-Narratives}. 



\end{abstract}

\section{Introduction}
\label{sec:intro}

Humans process information from their surrounding contexts from multiple modalities. These situated contexts are often derived from a modality (source) and expressed in another modality (target). Recent advances have seen a surge of interest in vision and language as source and target modalities respectively. One such widely studied task is image captioning \cite{hossain2019comprehensive, liu2019survey} which provides a textual description  $T$ given an image $I$. In contrast, visual storytelling \cite{huang2016visual} is the task of generating a sequence of textual descriptions ($\{ T_{1}, T_{2}, ... , T_{n} \} $) from a sequence of images ($\{ I_{1}, I_{2}, ... , I_{n} \} $). This sequential context is the differentiating factor in generation of visual narratives in comparison to image captioning in isolation.  This long form generation comprises of a coherent sequence of multiple sentences.

\begin{figure}[t!]
\centering
\includegraphics[trim=2.8cm 3.5cm 0cm 0cm,width=0.99\linewidth]{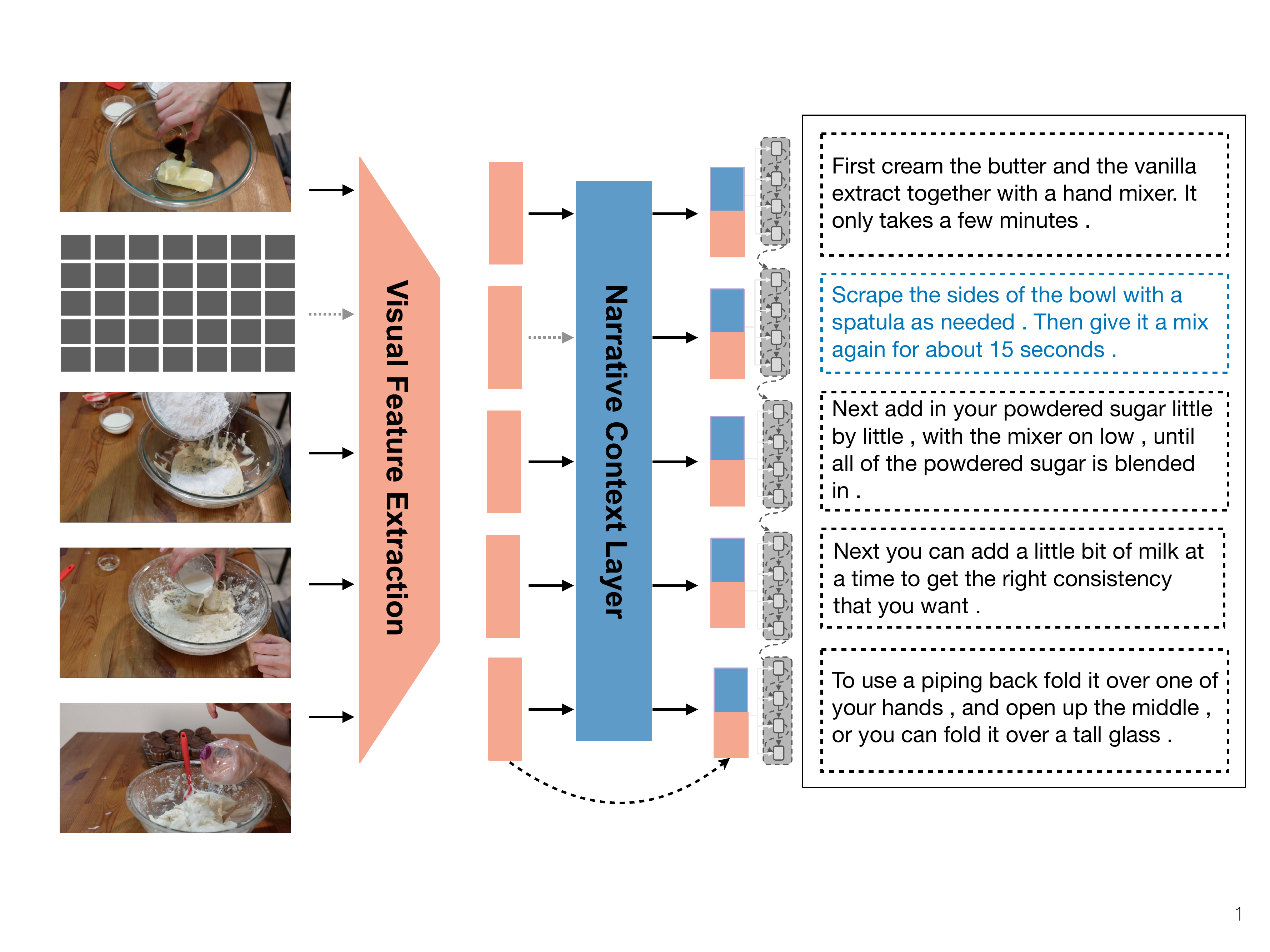}
\caption{ \small{Overview of infilling in visual procedures. Image in the second step is masked while the model generates the corresponding textual description from surrounding context.
}}
\label{fig:infilling_overview}
\end{figure}

A fundamental incongruity between how humans process information from multiple modalities and how we teach machines to do the same is that, humans are capable of bridging the information gap from surrounding contexts. Our training procedures do not take care of accommodating the same ability in a supervised learning paradigm.
Traditionally, the problem of missing context in long text generation is addressed using additional input such as entities, actions,  etc.,
\cite{DBLP:conf/acl/FanLD19, dong2019induction}, latent templates, external knowledge etc,. These are explicit methods to inject content during generation.
In contrast, in the spirit of simplicity, we propose infilling techniques to implicitly interpolate the gap between surrounding contexts from a stream of images. The training procedure incorporates masked contexts with the objective of a masked span prediction. We focus on two kinds of visual narratives namely, stories and procedures. We curated a large scale \textit{ViPT} dataset with pairwise image and text descriptions comprising of 46k procedures and 340k images. The percentage of unique words in each step in comparison to the rest of the recipe is about 60\% for ViST and 39\% for ViPT. This implies that overlapping contexts are predominant in procedures than stories datasets. This is usually because stories are more creative and diverse while procedures are in-domain. For both these reasons, we hypothesize that infilling technique is more effective in scenarios where it can leverage the vast context from the surrounding information to filling the missing pieces. 

To this end, we present our infilling based model to perform visual narrative generation and compare its effects on visual stories and procedures. The overview of the infilling based training procedure is presented in Figure \ref{fig:infilling_overview}. We conclusively observe that it is more effective in procedural texts with stronger contextual dependencies. We also present the effects of infilling during training and inference phases, and observe that infilling shows benefits during inference as well. Similarly, the infilling based techniques are also capable of generating longer sentences.
Interpolating contexts to generate narrative descriptions has potential applications in fields such as digital education \cite{hollingshead2018designing}, social media content \cite{gella2018dataset}, augmented reality \cite{dudley2018fast}, video games  \cite{kurihara2019automatic, ammanabrolu2019toward}, etc,. 
\noindent The main contributions of this paper are:
\begin{itemize}
\item We present a Visual Procedure Telling (ViPT) dataset similar to the Visual Storytelling (ViST) dataset with 46k procedures on various domains.
\item We demonstrate the efficacy of our visual infilling technique on narratives that have stronger contextual dependencies on the rest of the sentences. 
\end{itemize}

\section{Related Work}

\paragraph{Multimodal Language:} Language generation from visual modality has seen a steep rise in interest with the introduction of several large scale tasks such as image captioning \cite{hossain2019comprehensive}, visual question answering \cite{antol2015vqa} and visual dialog \cite{das2017visual, mostafazadeh2017image, de2017guesswhat}. While the task of generating a sentence from a single image i.e., image captioning has been well studied in the literature, generating a long form sequence of sentences from a sequence of images has been catching attention only in the recent past. Hence, the natural next step here is towards long form sequential generation in the form of stories, procedures etc., visual narrative telling.

\paragraph{Visual Storytelling: } \citet{huang2016visual} ventured into sequential step wise generation of stories by introducing visual storytelling (ViST). Recent methods have tackled ViST using adversarial learning, reinforcement learning \cite{wang2018no, huang2019hierarchically, hu2019makes}, modality-fusion \cite{smilevski2018stories}, traditional seq2seq models \cite{kim2018glac, jung2020hide, hsu2018using} and explicit structures \cite{DBLP:conf/acl/BosselutCWHC16, DBLP:conf/naacl/BiskBPC19}. \citet{chandu2019storyboarding} also proposed a dataset of 16k recipes in a similar form. While these are all cooking recipes, the ViPT dataset comprises a mixture of ten different domains. Also, our dataset is aboout 2.8 times larger than the storyboarding dataset with almost double the number of procedures in the domain of cooking recipes itself. Though the stories in ViST demonstrate a sense of continuity, the overarching sequential context is feeble. Procedures such as cooking recipes \cite{DBLP:conf/cvpr/SalvadorDNR19, DBLP:conf/cvpr/WangSLLH19} on the other hand, demonstrate this characteristic inviolably. This ensures a coherent underlying context and structure in the narrative. Hence, we present a large scale ViPT dataset to encourage research in this direction.


\begin{table*}[tbh]
\centering
\resizebox{\textwidth}{!}{%
\begin{tabular}{l|l|llllllllll}
\hline
\hline
\textbf{Dataset}    & \multicolumn{1}{c|}{\textbf{ViST}} & \multicolumn{10}{c}{\textbf{Visual Procedure Telling (ViPT)}}    \\ \hline
\textbf{Categories} & \textbf{stories}                   & \textbf{recipes} & \textbf{crafts} & \textbf{outdoors} & \textbf{lifestyle} & \textbf{technology} & \textbf{styling} & \textbf{fitness} & \textbf{hobbies} & \textbf{pets} & \textbf{misc} \\ \hline
\#narratives        & 50,136                             & 34,138           & 660             & 1,831             & 1,824              & 1,660               & 1,585            & 911              & 1,701            & 858           & 1,032         \\ 
\#images or steps   & 209,651                            & 203,519          & 8,658           & 20,526            & 20,959             & 19,221              & 18,112           & 9,935             & 19,145           & 9,599          & 11,853        \\ 
avg \#steps         & 5.00                               & 5.96             & 13.12           & 11.21             & 11.49              & 11.57               & 11.42            & 10.90            & 11.25            & 11.18         & 11.48         \\ 
avg \#words/step    & 11.35                              & 79.19            & 47.99           & 35.52             & 32.58              & 27.90               & 17.31            & 17.54            & 17.54            & 17.24         & 57.45         \\ \hline \hline
\end{tabular}%
}
\caption{Details of the ViST and \textit{Visual Procedure Telling} Dataset broken down into 10 categories}
\label{tab:dataset}
\end{table*}

\paragraph{Infilling and Masking: } The idea is motivated by cloze tasks \cite{taylor1953cloze} that addresses readability and understanding. However, recent advances in learning a masked language model \cite{devlin2019bert} paved way for a new trend in exploring masked contexts \cite{song2019mass, lewis2019bart}. 
Generation of meaning patches with missing portions of text is experimented by \citet{zhu2019text, donahue2020enabling, fedus2018maskgan} to generate meaningful patches.
Similarly, \citet{ippolito2019unsupervised} proposed a hierarchical model to generate middle span using a bag of predicted words from left and right contexts.  In a similar spirit, this paper studies the effects of infilling techniques for visual narrative generation. An alternate stream of work to improve the context in stories include providing supporting information such as entities \cite{clark2018neural, xu2018skeleton}, latent templates \cite{wiseman2018learning}, knowledge graphs \cite{yang2019knowledgeable}, etc., explicitly. In contrast to this, infilling provides an opportune platform to implicitly learn the contextual information.
Our work is positioned in the intersection of infilling and multimodal language generation.




\section{ViPT Description}
\label{sec:data}

While there are several types of narratives such as literary, factual and persuasive, this paper looks into stories and procedures. This section describes our new ViPT dataset and highlights the differences with ViST.

\paragraph{Procedures vs Stories: } 
Long form narratives are often characterized by three crucial properties: content, structure and surface form realization \cite{gatt2018survey}. 
Narrative properties such as content and structure in these forms are sufficiently contrastive between stories and procedures. Content in stories include characters and events while procedures include ingredients, materials and actions. 
Coming to the structure, stories typically start by setting a scene and the era followed by characterizing the participants and culminating with a solution if an obstacle is encountered. In contrast, a procedural text is often goal oriented and thereby typically begins by listing the ingredients/materials needed followed by a step by step description to arrive at the final goal. While stories can be metaphoric, sarcastic and humorous in surface realization, the sentences in procedures are often in imperative or instructional tone. 



\paragraph{2. Data Collection Process: } We manually examined around 10 blogging websites with various user written text on several \textit{how-to} activities. Among these we found that \textit{snapguide} and \textit{instructables} are consistent in the form of pairs of textual descriptions along with their images. We are going to release the scripts used to collect this data as well as preprocess them. We removed all the procedures in which atleast one image in each step is absent. Once all this preprocessing is done, the data contained the following categories in both the websites. These categories are based on the tags given by the bloggers to the articles they have written from among the categories that each website offers. These categories for each of these websites are:

\begin{itemize}
    \item \textit{snapguide: } recipes, games-tricks, sports-fitness, gardening, style, lifestyle, outdoors, beauty, arts-crafts, home, music, photography, pets, automotive, technology
    \item \textit{instructables: } crafts, cooking, teachers, circuits, living, workshop, outside
\end{itemize}

In union, they are a total of 18 categories. We manually examined a few procedures in each of the categories and regrouped them into 10 broad categories that are presented in Table \ref{tab:dataset}. A list of urls corresponding to the data is submitted along with the paper.

\begin{table}[tbh]
\centering
\resizebox{0.4\textwidth}{!}{%
\begin{tabular}{l|l|l}
\hline
\hline
\textbf{category}    & snapguide       & instructables            \\
\hline
\textbf{recipes}    & desserts,          & cooking            \\
\textbf{}    &  food         &             \\
\textbf{crafts}     & arts-crafts            & craft              \\
\textbf{outdoors}   & outdoors,     & outside            \\
\textbf{}   & gardening    &             \\
\textbf{lifestyle}  & lifestyle,         & living             \\
\textbf{}  &  home        &              \\
\textbf{technology} & technology,  & circuits           \\
\textbf{} & automotive &            \\
\textbf{styling}    & style,           &                    \\
\textbf{}    &  beauty          &                    \\
\textbf{fitness}    & sports-fitness         &                    \\
\textbf{hobbies}    & music,      &                    \\
\textbf{}    &  photography     &                    \\
\textbf{pets}       & pets                   &                    \\
\textbf{misc}       & games-tricks           & teachers, \\
\textbf{}       & games-tricks           & workshop\\
\hline
\hline
\end{tabular}%
}
\caption{Regrouping the categories in ViPT dataset}
\label{tab:my-table}
\end{table}

\paragraph{Visualization of topics: }
Each of the categories in our Visual Procedure Telling (ViPT) are analyzed for the topics present in them. 
To get a more detailed understanding of these topics in the dataset, we hosted the topic visualizations here: \href{https://visual-narratives.github.io/Visual-Narratives/}{visual-narratives.github.io/Visual-Narratives/}.

\paragraph{ViPT dataset: } Though stories have the potential to exhibit the properties listed above, it is challenging to observe them in the ViST dataset \cite{huang2016visual} owing to the shorter sequence lengths. The extent to which adjacent groups of sentences have overlapping contexts is high in procedures as compared to stories. We had previously gathered cooking recipes to experimentally demonstrate a scaffolding technique to improve structure in long form narratives \citet{chandu2019storyboarding}. We extend this work to gather procedures or \textit{`how-to'} articles that have step by step instructions along with an associated pairwise image to each step in several domains.
To facilitate multi-domain research with stronger interleaved contexts between surrounding steps, we present a large scale \textit{visual procedure telling} dataset with 46k procedures comprising of 340k pairwise images and textual descriptions. It is carefully curated from a number of \textit{how-to} blogging websites. Our dataset comprises of pairwise images and textual descriptions of the corresponding images, typically describing a step in a procedure. This means that each description of the step is tethered to an image. This makes it a \textit{visual narrative telling} task.
We categorized the dataset into 10 distinct domains including recipes, crafts, outdoors, lifestyle, technology, styling, fitness, hobbies, pets and miscellaneous. The category wise details of the dataset are presented in Table \ref{tab:dataset}. As we can observe, the dataset is domainated by cooking recipes which are relatively of similar sizes with ViST compared to the rest of the domains. 

\paragraph{Differences between ViPT and ViST datasets: } As observed in Table \ref{tab:dataset}, the average number of steps in ViPT is higher than that of ViST. However, the average number of steps in recipes and stories are similar which is 5.96 and 5.00 respectively. The average number of words per step in ViPT is also much higher, thereby presenting a more challenging long form text generation task. Despite the average number of steps being similar, the average length of each step i.e, the number of words per step in cooking recipes is about 7 times that of stories. Typically, each step in the ViPT dataset comprises of multiple sentences that is indicative of the corresponding image. This is as opposed to ViST dataset, which has a single sentence per step.
These long sequences also present a case for dealing with larger vocabularies as well. 
The recipes category alone has a vocabulary of 109k tokens while the same for stories is 25k. We also compared the diversity in vocabulary of each step by computing the average percentage of unique words in a step with respect to the rest of the narrative. While this number is a high 60\% for ViST, it is 39\% for ViPT. This means that there are about 40\% of the words in each step in ViST that are overlapping with the rest of the story. This could be owed to the way the dataset is gathered by asking the annotators to pick a sequence of images that are likely to make a coherent story and then describing these images in sequence. While the stories-in-sequences sufficiently distinguish themselves from descriptions-in-isolation, the overlapping contexts are not high compared to procedures. The overlapping contexts for procedures is about 61\%. This reveals the stronger cohesive and overlapping contexts in the ViPT dataset, as compared to the ViST datasets. These overlapping contexts motivates the idea of generating a sentence by bridging the contexts from surrounding sentences. Hence it forms a suitable test bed to learn interpolation from surrounding contexts 
with infilling technique.


\begin{table*}[tbh]
\small
\centering
\begin{tabular}{l|l|l|l|l|l|l|l|l}
\hline
\hline
\textbf{Dataset} & \multicolumn{4}{c|}{\textbf{Stories}}     & \multicolumn{4}{c}{\textbf{Recipes}}          \\ \hline
\textbf{Model}            & \textbf{XE}    & \textbf{V-Infill}    & \textbf{V-InfillR} & \textbf{INet}            & \textbf{XE}    & \textbf{V-Infill}             & \textbf{V-InfillR}       & \textbf{INet}  \\ \hline
BLEU-1           & 62.05 & 61.58 & 61.84    & \textbf{63.31}  & 28.61 & \textbf{29.73} & 28.61          & 25.10 \\ 
BLEU-2           & 38.31 & 37.27 & 37.81    & \textbf{39.60}  & 16.89 & \textbf{17.50} & 17.01          & 13.36 \\ 
BLEU-3           & 22.68 & 21.70 & 22.42    & \textbf{23.62}  & 10.50 & \textbf{10.83} & 10.59          & 6.51  \\ 
BLEU-4           & 13.74 & 12.96 & 13.69    & \textbf{14.30} & 5.68  & \textbf{5.81}  & 5.71           & 3.60  \\ 
METEOR           & 35.01 & 34.53 & 35.08    & \textbf{35.57}  & 26.72  & 27.26          & \textbf{27.51} & 25.62 \\ 
ROUGE\_L         & 29.66 & 29.12 & 29.65    & \textbf{30.14}  & 21.64 & \textbf{22.02} & 18.66          & 20.43 \\ \hline \hline
\end{tabular}
\caption{Performance of different models on stories (from ViST) and recipes (from ViPT) datasets}
\label{tab:results_overall}
\end{table*}

\begin{table*}[tbh]
\centering
\small
\begin{tabular}{l|l|l|l|l|l|l|l|l|l|l|l|l}
\hline
\hline
\textbf{Infill Index} & \multicolumn{2}{c|}{\textbf{0}}        & \multicolumn{2}{c|}{\textbf{1}} & \multicolumn{2}{c|}{\textbf{2}} & \multicolumn{2}{c|}{\textbf{3}} & \multicolumn{2}{c|}{\textbf{4}} & \multicolumn{2}{c}{\textbf{5}}        \\ \hline
\textbf{Model}        & \textbf{XE}            & \textbf{V-Infill}            & \textbf{XE}         & \textbf{V-Infill}        & \textbf{XE}         & \textbf{V-Infill}        & \textbf{XE}         & \textbf{V-Infill}        & \textbf{XE}         & \textbf{V-Infill}        & \textbf{XE}            & \textbf{V-Infill}            \\ \hline
BLEU-1           & 20.9          & 29.7          & 22.8       & 29.8      & 23.5       & 29.9      & 24.4       & 30.4      & 25.5       & 31.0      & \textbf{26.4} & \textbf{31.5} \\ 
BLEU-2           & 12.5          & 18.0          & 13.2       & 17.6      & 13.6       & 17.5      & 14.2       & 17.8      & 14.9       & 18.2      & \textbf{15.4} & \textbf{18.6} \\ 
BLEU-3           & 7.9           & 11.1          & 8.2        & 10.7      & 8.4        & 10.8      & 8.8        & 10.9      & 9.2        & 11.1      & \textbf{9.6}  & \textbf{11.4} \\ 
BLEU-4           & 4.2           & 5.8           & 4.2        & 5.6       & 4.4        & 5.6       & 4.7        & 5.7       & 4.9        & 5.8       & \textbf{5.1}  & \textbf{6.0}  \\ 
METEOR       & 27.6 & 27.8 & 26.4       & 27.1      & 26.0       & 26.9      & 26.3       & 27.1      & 26.6       & 27.2      & 26.8          & 27.4          \\
ROUGE\_L     & 20.9          & 22.4 & 20.3       & 21.8      & 20.6       & 21.8      & 21.0       & 21.9      & 21.3       & 22.0      & \textbf{21.5} & 22.1          \\ \hline \hline
\end{tabular}
\caption{Performance of infilling during inference for recipes in Visual Procedure Telling}
\label{tab:results_recipe_test}
\end{table*}

\begin{table*}[tbh]
\small
\centering
\begin{tabular}{l|l|l|l|l|l|l|l|l|l|l}
\hline
\hline
\textbf{Infill Index} & \multicolumn{2}{c|}{\textbf{0}} & \multicolumn{2}{c|}{\textbf{1}} & \multicolumn{2}{c|}{\textbf{2}} & \multicolumn{2}{c|}{\textbf{3}} & \multicolumn{2}{c}{\textbf{4}} \\ \hline
\textbf{Model}        & \textbf{XE}    & \textbf{V-Infill}             & \textbf{XE}         & \textbf{V-Infill}        & \textbf{XE}        & \textbf{V-Infill}       & \textbf{XE}         & \textbf{V-Infill}        & \textbf{XE}             & \textbf{V-Infill}  \\ \hline
BLEU-1           & 60.9  & 63.0  & 60.8       & 62.0      & 60.3       & 61.9      & 60.5       & 62.2      & \textbf{61.8}  & \textbf{63.3}  \\ 
BLEU-2           & 37.0  & 39.5  & 36.9       & 38.6      & 37.0       & 38.4      & 37.0       & 38.7      & \textbf{38.1}  & \textbf{39.6}  \\
BLEU-3           & 21.7  & 23.7  & 21.6       & 23.1      & 21.8       & 22.9      & 21.8       & 23.2      & \textbf{22.5}  & \textbf{23.7}  \\ 
BLEU-4           & 13.1  & 14.4  & 13.1       & 14.1      & 13.2       & 13.9      & 13.3       & 14.3      & \textbf{13.8}  & \textbf{14.5}  \\ 
METEOR       & 34.9  & 35.4  & 34.8       & 35.1      & 35.2       & 35.2      & 35.1       & 35.3      & \textbf{35.2}  & \textbf{35.5}  \\ 
ROUGE\_L       & 29.3  & 30.2  & 29.2       & 29.9      & 29.1       & 30.0      & 29.2       & 30.0      & \textbf{29.5}  & \textbf{30.3}  \\ \hline \hline
\end{tabular}
\caption{Performance of infilling during inference for Visual Story Telling}
\label{tab:results_story_test}
\end{table*}

\section{Models Description}
\label{sec:models}
This section describes the baseline model and the infilling techniques adopted on top of it.

We present infilling based techniques for learning missing visual contexts to generate narrative text from a sequence of images. As the ViST and recipes category in ViPT are of comparable sizes (both in terms of data size and the average number of steps per instance), we perform comparative experimentation on these two categories. We leave experimenting with all the domains for our future work, especially learning from one domain to generate the sequences in other domains. For our ViPT category, we use 80\% for training, 10\% for validation and 10\% for testing. The stories are composed of 5 steps and the cooking recipes are trucated to 5 steps to perform a fair comparison of the effect of the index being infilled. An overview of infilling based training is depicted in Figure \ref{fig:infilling_overview}. 
The underlying encoding and decoding stages are described here. 

\paragraph{Encoding: } Models 1, 2 and 3 here show different variants of encoding with and without infilling. Model 4 is the state of the art model for generating stories on ViST. Note that the encoding part of the missing contexts varies between these models while the decoding strategy remains the same to compare (i) the performance of encoding masked contexts as opposed to not masking, and (ii) the performance of masked span prediction between stories and procedures.

\paragraph{1. XE (baseline): } We choose a strong performing baseline model based on sequence to sequence modeling with cross entropy (XE) loss inspired from \citet{wang2018no}. It is a CNN-RNN architecture. The visual features are extracted from the penultimate layer of ResNet-152 by passing the resized images ($\{ I_{1}, I_{2}, ... , I_{n} \} $) of size 224 X 224. These represent the image specific local features ($\{ l_{1}, l_{2}, ... , l_{n} \} $). These features are then passed through a bidirectional GRU layer to attain narrative level global features ($\{ g_{1}, g_{2}, ... , g_{n} \} $) constituting the narrative context layer in Figure \ref{fig:infilling_overview}. 

\paragraph{2. V-Infill: } We introduce an infilling indicator function on the underlying XE model by randomly sampling an infilling index ($in_{idx}$). This is used to construct the final infilled local features as follows.

\[
l_k \small{ (\forall k, s.t. 0< k\leq n) } = \begin{cases}
zero\_tensor &\text{if k=$in_{idx}$}\\
l_k &\text{otherwise}
\end{cases}
\]

Other than the sampled $in_{idx}$, the rest of the local features for other indices remain the same. The local features for $in_{idx}$ are all masked to a zero tensor. 
The dropout of an entire set of local features from an image forces the model to learn to bridge the context from the left and the right images of $in_{idx}$. The model is optimized to predict the rest of the steps where images are present along with the \textit{masked span prediction}. In this way, the infilling mechanism encourages our underlying seq2seq model to learn the local representation of the missing context from contextual global features in the narrative context layer.



\paragraph{3. V-InfillR: } This model varies the \textbf{R}ates in which local features are masked as training proceeds based on the indicator function above in the V-Infill model. Scheduling the number of missing features itself is a hyperparameter and we used the following setting. In the first quarter of training epochs, none are masked, then increasing it to 1 local feature for the next quarter and leaving it at 2 for the last two quarters. This is similar to the settings observed in INet model. We have experimented with other settings of scheduling but this one performed better than the others.

As mentioned earlier, the encoding of the local features change based on the infilling technique being used in each of the above strategy. As we can see, the contribution of the global features to reconstruct the local missing context is intuitively expected to perform well in the case of narratives with overlapping contexts. Hence, we hypothesize that the infilling technique that interpolates between steps that constitute words or phrases that are similar to those of the surrounding steps benefit from this technique. A \textit{`how-to'} style of narrative explaining a procedure is more in-domain as compared to the stories and hence hypothesize that our infilling based encoding approaches perform relatively better on procedures.
We then use the encoded representation to decode each step of the procedure or story. The decoding strategy is explained next which is the same in all the three of the aforementioned models.


\paragraph{Decoding: } In all the above models, $g_k$ are fed into a GRU decoder to predict each word ($\boldsymbol{\hat{w}}_{t}$) of the step (k). The same is done for generating each step in the five steps. In the infilling methods, the decoding strategy is agnostic to the missing context in the local features. The global features that bridges the contexts in the encoding is used directly as input to the decoder. In other words, the network remains the same once the global features are predicted. We perform beam search with a beam size of 3 during inference. Here $\tau$ is the number of words in each step and $t$ is the current time step.
\[
\boldsymbol{\hat{w}}_{t} \sim \prod_\tau Pr(\boldsymbol{\hat{w}}_t^{\tau} | \boldsymbol{\hat{w}}_t^{<{\tau}}, \boldsymbol{g}_k)
\]

\paragraph{4. INet: } We re-implemented the model achieving the state of the art results \cite{hu2019makes} on the visual storytelling dataset. Additionally, they use a relational embedding layer that captures relations across spatio-temporal sub-spaces. Our replication of their model is close to the scores reported in their paper, though not exact. Our re-implementation achieved a 35.5 METEOR and 63.3 BLEU-1 in comparison to the scores reported in their paper which are 35.6 and 64.4. 

\paragraph{Hyperparameter Setup:} We use a GRU with hidden dimension of 256 for encoder and 512 for decoder. The word embedding dimension is 512. The learning rate is 4e-4 optimized with Adam and smoothing of 1e-8. We use a dropout of 0.2 and momentum of 0.9 with a gradient clipping of 10. The performance when experimented with a transformer based encoder along with autoregressive decoding is comparatively lesser and hence we proceed with a GRU based model. Based on the average number of steps in recipes from Table \ref{tab:dataset} which is 5.96, we truncate the recipes to 6 steps.

\section{Results and Discussion}
\label{sec:results}

\begin{figure*}[tbh]
\centering
\includegraphics[trim=0cm 1cm 0cm 0cm,width=0.99\linewidth]{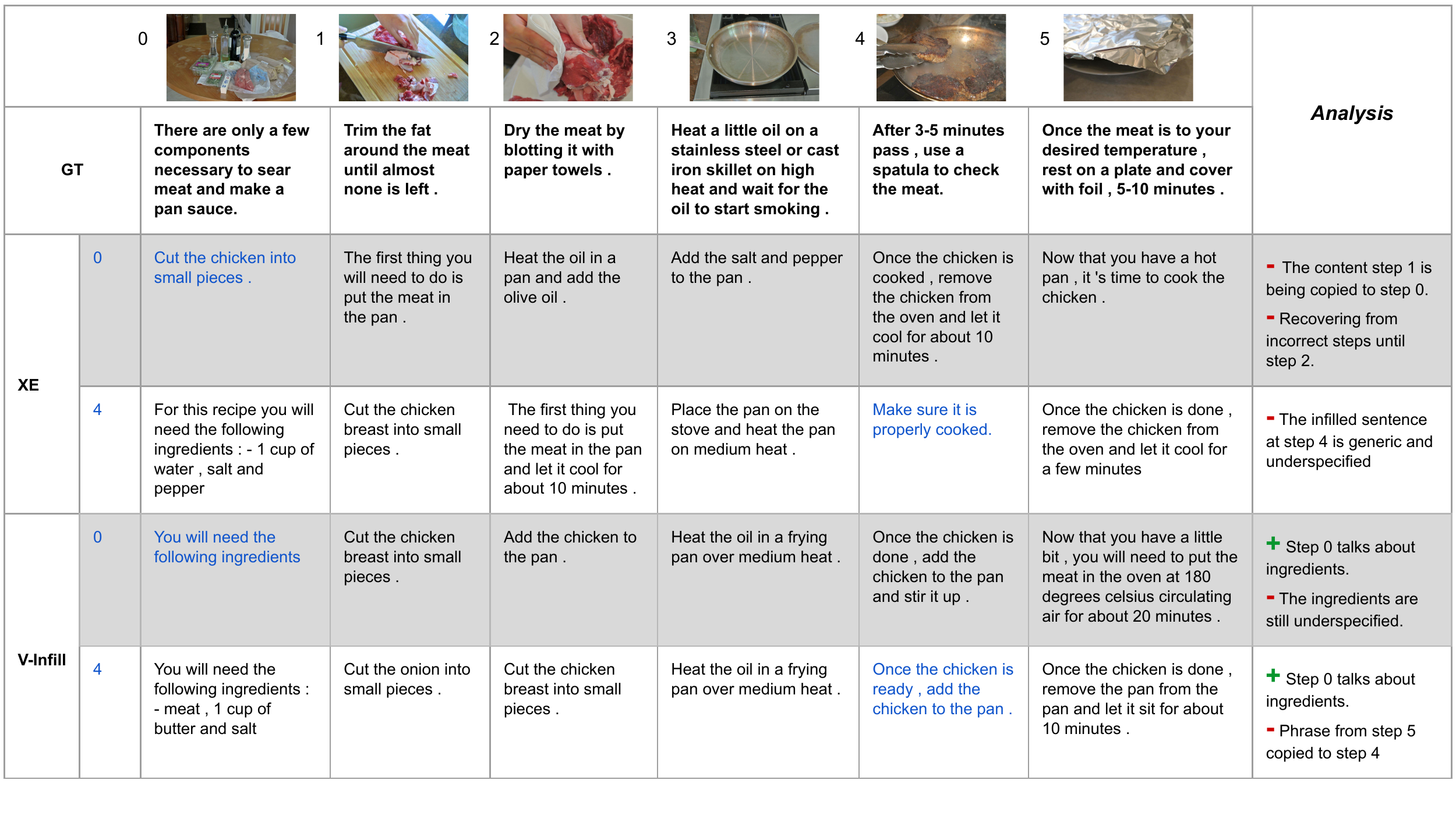}
\caption{Comparison of V-Infill and XE dealing with infilling context during inference (for making \textit{chicken roast}). GT corresponds to the ground truth step. The index in each row corresponds to the index of the missing image. }
\label{fig:analysis}
\end{figure*}

In this section, we present the effects of infilling both during both training and inference on ViST and ViPT datasets. We also present an analysis based on the length of generated sequences along with a qualitative demoonstration.

\paragraph{Infilling during training: } The overall performance of the models is presented in Table \ref{tab:results_overall}. Both the infilling model variants achieve higher scores on the recipes while not decreasing their performances on stories. We also observed that increasing the number of masked local features beyond 2 drastically decreases the performance on both datasets.


\paragraph{Infilling during inference for Visual Procedure Telling: } Acquiring parallel pairwise image and narrative data in the wild is often not feasible. Hence, we perform infilling not only at train time but also at inference time to evaluate the ability of the model to bridge contexts when the corresponding image is absent and deal with real world data imputation scenarios. Table \ref{tab:results_recipe_test} demonstrate the performance of the V-Infill model in comparison with the XE model when different indices are infilled during inference stage. As observed, the automatic scores get affected detrimentally when the infilled index is to the left, i.e., a lower index. This is because usually the beginning of the sentence comprises of introducing the dish followed by listing down the ingredients. For this reason, the density of the number of entities that are present in the beginning of the procedure is usually higher. Hence reconstructing that from the rest of the recipe is difficult. However, as we move from left to right, i.e as we gradually increase the infilled index, we observe an increasing trend in the automatic metric.

\paragraph{Infilling during inference for Visual Story Telling:} Table \ref{tab:results_story_test} demonstrates the effects of infilling various indices during inference. This table is analogous to Table \ref{tab:results_recipe_test} for stories. As we can see, a similar trend in the increase in all the automatic metrics are present as we move the infill index to the right of the story. While that is still the case, a very interesting observation is that the difference between the performance of XE and Infill models for any given index is much higher for recipes compared to stories. The infilling technique is bringing much more value to the task when the nature of the text is procedural and dependent more on the surrounding contexts.

\paragraph{Lengths of generated sequences :}  We compare infilling during inference between baseline XE model and our V-Infill model in Table \ref{tab:results_recipe_test}. While the METEOR scores remain comparable, the BLEU scores steadily increase as we move the $in_{idx}$ to the right. Specifically, these jumps are bigger after step 3. Quantitatively, this is the result of the model being able to produce longer sequences as we move to the right as BLEU gets penalized for short sentences. Qualitatively, this implies that the initial steps like specifying the ingredients are more crucial as compared to later ones. A similar observation emerges by analyzing the effects of infilling during training. The average length of generated recipes by XE is 71.26 and by V-Infill is 76.49. A similar trend is observed for stories in Table \ref{tab:results_story_test}.


\paragraph{Qualitative Discussion: } Figure \ref{fig:analysis} demonstrates an example of generated samples by infilling different indices. The top row shows the steps in the ground truth steps for the corresponding images. The indices on top row are the indices of the images or the steps and the indices on the left column (in blue) are the indices whose local features are masked. As observed, the XE model depicts two strategies to recover the missing context. The first is copying the contents that are similar from the adjacent step directly. For instance, while the $0^{th}$ index of the image is masked, the XE model generates \textit{cutting} from \textit{trimming} and \textit{chicken} from \textit{meat} from the following step. This has nothing to do with the actual description of the corresponding step. However our V-Infill model is able to generate the sentence depicting that it is listing ingredients in this case. Since the first step is incorrectly generated by the baseline, it makes it harder to recover and generate the correct sequence for the rest of the procedure. The second is the strategy of generating generic sentences. When infilled index is at 4, the baseline model generates a sentence that is generic and not specific to the given set of images. In this case, it generates a statement that says to make sure that it is properly cooked. Our V-Infill model is able to bridge the context from step 3 about heating the oil and step 5 about removing the pan and hence interpolates the missing context to be placing the chicken on the pan.

Despite the recovering strategies used in both these methods, there is a common problem observed in the generated steps. The details in the steps are omitted thereby leading to the problem of \textit{under-specification}. For instance, the actions in step 4 are under-specified by XE when the infilled index is 4. Similarly the V-Infill model under-specifies the ingredients

\begin{figure}[tbh]
\centering
\includegraphics[trim=0cm 1cm 0cm 1cm,width=0.99\linewidth]{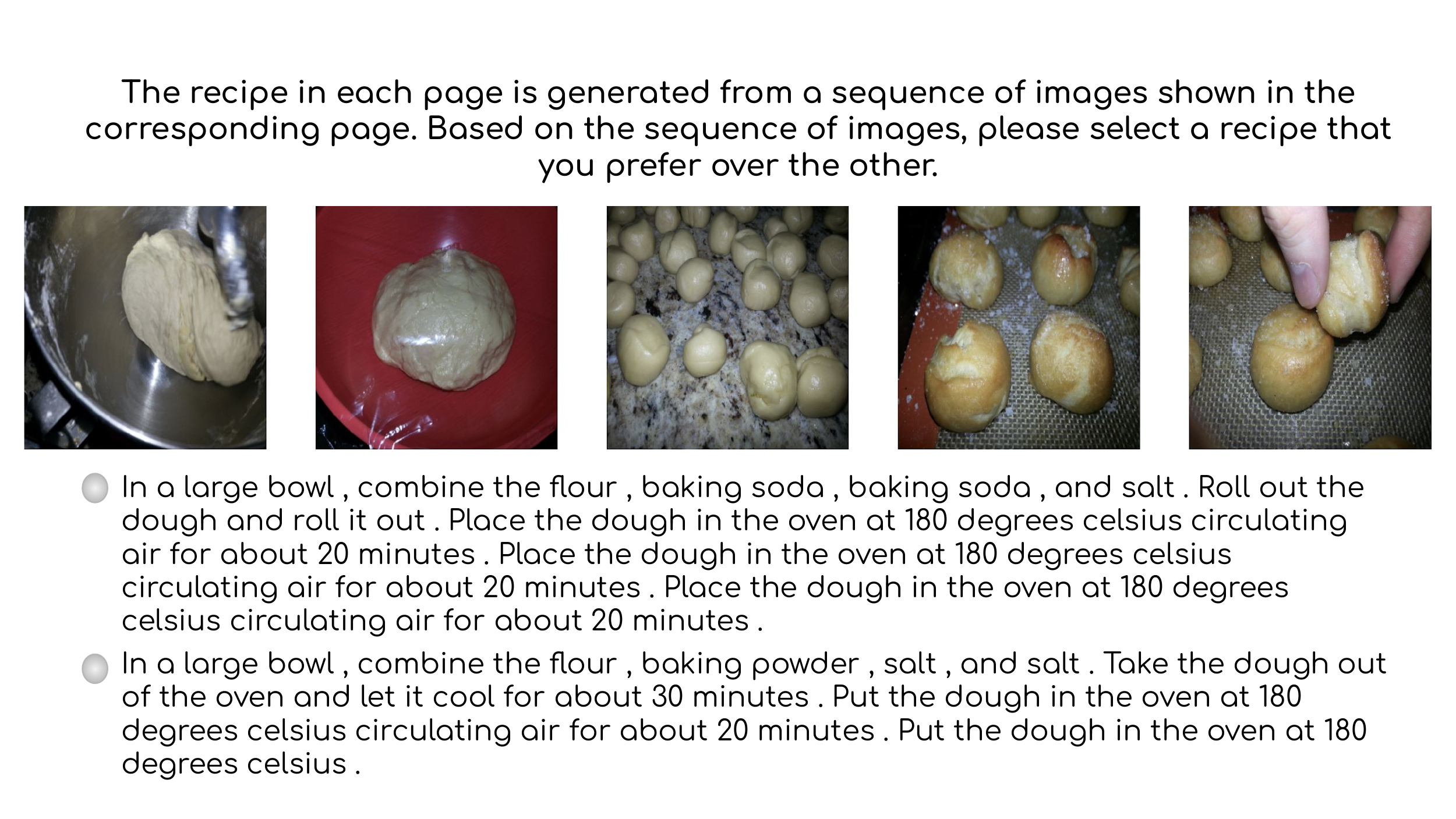}
\caption{Human Evaluation Interface for an example of generated recipes with both techniques.}
\label{fig:human_eval}
\end{figure}

\paragraph{Human Evaluation: } 
Figure \ref{fig:human_eval} depicts a screenshot of our human evaluation interface. A sequence of images are presented on top of the screen. This evaluation is conducted to compare between XE and V-Infill model. The generated sentences from both the models, in this case XE and V-Infill are presented after the images. Note that the generated outputs are presented in arbitrarily random order for each example to ensure there is no bias while performing preference testing. Human subjects are asked to pick one of the generated recipes for the given sequence of images based on the relevance to them. 10 such recipes are presented for each user and we avergaed the preference scores among 20 evaluators.


\section{Conclusions and Future Work}

We demonstrate that infilling is a simple yet effective technique and a step towards maximizing the utilization of surrounding contexts in visual narratives. Infilling is the strategy of enabling the model to learn surrounding contextual information by masking spans of input while the decoding attempts in generating the entire text. The input to the model is provided with masked contexts and the model is optimized with the objective of masked span prediction. We hypothesize that this technique provides gains in narratives with higher extent of overlapping contexts, since this provides an opportunity to reconstruct the missing local context from the overall global context. To this end, we introduce a new large scale ViPT dataset of 46k procedures and 340k image-text pairs comprising 10 categories. 
To experimentally support our hypothesis,  we compare the performance of our model. We conclusively show the higher significance of infilling based techniques in visual procedures compared to visual stories. We also perform comparisons between the infilling during training and inference phases. With infilling during training, our V-Infill model performs better on visual procedures in comparison to stories. With infilling during inference, our v-infill model performs better on both stories and procedures. In the case of stories, infilling during inference is surprisingly better than fully supervised seq2seq model and very close the state of the art model as well. 
In future, we plan to explore the following two directions: (1) interpolating the contexts between consecutive steps by introducing a new infilled image; this addresses the data imputation problem as well as generating longer explanations to unclear steps. And (2) addressing the underspecification problem by controlling the content in infilled image with explicit guidance; this is as opposed to the implicit content filling that we perform throough interpolation. These infilling techniques are also immensely useful when dealing with data imputation with missing contexts and collaborative authoring in real world scenarios.



\bibliographystyle{acl_natbib}
\bibliography{anthology,emnlp2020}

\end{document}